\documentclass[journal,onecolumn]{IEEEtran}

\usepackage{amsmath,graphicx}
\usepackage{algorithm}
\usepackage[noend]{algpseudocode}
\usepackage{stfloats}
\usepackage{xcolor}
\usepackage{amsfonts}
\usepackage{amssymb}
\usepackage{cite}
\usepackage{array}
\usepackage{subcaption}
\usepackage{times}
\usepackage{epsfig}
\usepackage{latexsym}
\usepackage{epstopdf}
\usepackage{verbatim}
\usepackage{units}
\usepackage{amsthm}
\usepackage{placeins}
\usepackage{afterpage}
\usepackage{dsfont}
\usepackage{soul}
 \usepackage{bbm}
\usepackage{multicol}
\usepackage{multirow}
\usepackage{mathtools}
\usepackage[cmintegrals]{newtxmath}
\usepackage{url}
\newcolumntype{P}[1]{>{\centering\arraybackslash}p{#1}}
\newcolumntype{M}[1]{>{\centering\arraybackslash}m{#1}}

\newtheorem{definition}{Definition}

\ifCLASSINFOpdf

\else

\fi

\begin{document}
%
\title{Collaborative Learning over Wireless Networks: An Introductory Overview}
%
%
%

\author{Emre~Ozfatura$^{a}$,
        Deniz~G{\"u}nd{\"u}z$^{b}$,
        and~ H. Vincent Poor$^{c}$
\thanks{$^{a}$Department of Electrical and Electronic Engineering,Imperial College London, London, UK}
\thanks{$^{b}$Department of Electrical and Electronic Engineering,Imperial College London, London, UK} 
\thanks{$^{c}$Department of Electrical Engineering, Princeton University, Princeton, NJ}}


\maketitle

%
\IEEEpeerreviewmaketitle

The number of devices connected to the Internet has already surpassed 1 billion. With the increasing proliferation of mobile devices the amount of data collected and transmitted over wireless networks is growing at an exponential rate. Along with this growth in data traffic, its content is also rapidly changing thanks to new applications, such as autonomous systems, factory automation, and wearable technologies, referred to as the Internet of things (IoT) paradigm in general. While the current data traffic is dominated by video and voice content that are transmitted mainly for user consumption, data generated by IoT devices is intended for machine analysis and inference in order to incorporate intelligence into the underlying IoT applications. Current approach to IoT intelligence is to offload all the relevant data to a cloud server, and train a powerful machine learning (ML) model using all the available data and processing power. However, such a `centralized' solution is not applicable in many ML applications due to the significant communication latency it would introduce. Moreover, IoT devices are typically limited in power and bandwidth, and the communication links become a major bottleneck as the volume of collected data increases. For example, autonomous cars are expected to generate 5 to 20 terabytes of data each day. The increasing data volume becomes particularly challenging when the `information density' of the collected data is low, i.e., large volumes of data needs to be offloaded with only limited relevant information for the underlying learning task. Centralized approaches also incite privacy concerns, and users are increasingly averse to sharing their data with third parties even in return of certain utility. Privacy concerns can be particularly deterring for personal data, which is often the case with IoT devices, even if it may not be obvious at first. 

The solution to all these limitations is to bring the intelligence to the network edge by enabling wireless devices to implement learning algorithms in a distributed and collaborative manner \cite{Gunduz:CL:20, Park:PIEEE:19}. The two core components behind the recent success of ML algorithms are massive datasets and computational power, which allows training highly complex models. Both of these are abundant at the network edge, but in a highly distributed manner. Hence, a main challenge in edge intelligence is to design learning algorithms that can exploit these distributed resources in a seamless and efficient manner. 

In this chapter, we will mainly focus on collaborative training across wireless devices. Training a ML model is equivalent to solving an optimization problem, and many distributed optimization algorithms have been developed over the last decades. 
These distributed ML algorithms provide data locality; that is, a joint model can be trained collaboratively while the data available at each participating device remains local. This addresses, to some extend, the privacy concern. They also provide computational scalability as they allow exploiting computational resources distributed across many edge devices. However, in practice, this does not directly lead to a linear gain in the overall learning speed with the number of devices. This is partly due to the communication bottleneck limiting the overall computation speed.  Additionally, wireless devices are highly heterogeneous in their computational capabilities, and both their computation speed and communication rate can be highly time-varying due to physical factors. Therefore, distributed learning algorithms, particularly those to be implemented at the wireless network edge, must be carefully designed taking into account the impact of time-varying communication network as well as the heterogeneous and stochastic computation capabilities of devices.

In recent years, many solutions have been proposed in the ML literature to reduce the communication load of distributed ML algorithms to mitigate the communication bottleneck; however, these solutions typically aim at reducing the amount of information that needs to be exchanged between the computing servers, assuming rate-limited perfect communication links, and they do not take into account the physical characteristics of the communication medium. However, such an abstraction of the communication channel is far from reality, particularly in the case of wireless links. Converting the wireless channel into a reliable bit pipe may not even be possible (e.g., in the case of fading channels), or may introduce substantial complexity and delays into the system, and increase the energy cost. Moreover, such an approach is based on an inherent separation between the design of ML algorithms and the communication protocols enabling the exchange of messages between devices. 

Our goal in this chapter is to show that the speed and final performance of distributed learning techniques can be significantly improved by taking the wireless channel characteristics into account. We will provide several examples and point to some of the recent literature for further details. We will mainly focus on the joint optimization of the parameters of the underlying communication protocols, e.g., resource allocation, scheduling, etc., together with the underlying distributed learning framework. An alternative approach is to design the learning and communication algorithms jointly, completely stepping out of the current digital communication framework \cite{Gunduz:CL:20, Amiri:TSP:20, Amiri:TWC:20}. 

In the remainder of the chapter, we will first introduce the general collaborative learning problem. We will then present centralized and decentralized approaches to collaborative learning, and highlight federated learning (FL) as a popular distributed learning framework. Afterwards, we will overview some of the techniques that can be used to reduce the communication load of distributed learning. We will then present device scheduling and resource allocation algorithms for distributed learning over wireless networks. We should emphasize that the goal of this chapter is not to provide a comprehensive survey of all existing works on this fast developing research area, but to provide an introduction to the challenges and typical solution approaches. 

\section{Collaborative Learning} \label{s:collaborative_learning}

Many parameterized ML problems can be modeled as the following {\em stochastic optimization problem}:
\begin{equation}\label{problem}
\min_{\boldsymbol{\theta}\in\mathbb{R}^{d}}  \mathbb{E}_{\zeta \sim \mathcal{P}}F(\boldsymbol{\theta}, \zeta),
\end{equation}
where $\boldsymbol{\theta}\in\mathbb{R}^{d}$ denotes the model parameters to be optimized,  $\zeta$ is the random data sample with distribution $\mathcal{P}$, and $F$ is the problem specific stochastic loss function. Typically the underlying distribution $\mathcal{P}$ is not known, and instead we have access to a dataset $\mathcal{D}$ sampled from this distribution. Accordingly, we minimize the empirical loss function $\mathbb{E}_{\zeta \sim \mathcal{D}}F(\boldsymbol{\theta}, \zeta)$. This problem is often solved iteratively using stochastic gradient descent (SGD), where the model is updated at each iteration along the direction of the gradient estimate, computed using a subset of the dataset. That is, at iteration $t$, we update the model $\boldsymbol{\theta}_t$ as follows
\begin{equation}
    \boldsymbol{\theta}_{t+1} = \boldsymbol{\theta}_{t}  - \eta_t \cdot \mathbf{g}_t,
\end{equation}
where 
\begin{equation}\label{gradient_est}
    \mathbf{g}_t = \nabla_{\boldsymbol{\theta_t}} F(\boldsymbol{\theta}_t, \zeta_t),
\end{equation}
$\zeta_t$ is a random sample from the dataset $\mathcal{D}$, and $\eta_t$ is the learning rate at iteration $t$. Note that $\mathbf{g}_t$ is an unbiased estimate of the full gradient of $\mathbb{E}_{\zeta \sim \mathcal{D}}F(\boldsymbol{\theta}, \zeta)$. SGD is guaranteed to converge to the optimal solution under certain conditions on function $F$ (e.g., convex and smooth), and it is particularly beneficial in practice when the dataset is large, and hence, a full gradient computation at each iteration is costly. 

In the distributed setting, we consider $N$ devices denoted by set $\mathcal{S}$, each with its own local dataset, denoted by $\mathcal{D}_i$. Stochastic optimization problem given above can be solved in a distributed manner over these $N$ devices by rewriting the minimization problem in (\ref{problem}) as
\begin{equation}\label{DSO}
\min_{\boldsymbol{\theta}\in\mathbb{R}^{d}} f(\boldsymbol{\theta})= \frac{1}{N}\sum^{N}_{i=1}\underbrace{\mathbb{E}_{\zeta \sim \mathcal{D}_{i}}F(\boldsymbol{\theta},\zeta)}_{\mathrel{\mathop:}=f_{i}(\boldsymbol{\theta})},
\end{equation}
which is an average of $N$ stochastic functions. The main idea behind collaborative learning is to solve the minimization problem in (\ref{problem}) in a distributed manner, such that each device tries to minimize its local loss function $f_{i}(\boldsymbol{\theta})$, while seeking a consensus with other devices on the global model $\boldsymbol{\theta}$. The consensus is facilitated either  with the help of a central entity called parameter server (PS), or by communicating with only neighboring devices. Hence, based on the consensus framework, collaborative learning methods can be divided into two classes, {\em centralized} and {\em decentralized}. The centralized approach has received significantly more attention in the literature due to its simplicity in both the implementation and the analysis. Before going into the details of the solution techniques, we briefly introduce the distributed learning problem in both cases and highlight the main differences. 

\subsection{Centralized Learning}
We classify a distributed learning framework as centralized when the learning framework is orchestrated by a PS. To be more precise, in the centralized framework each user/worker performs local computation under the supervision of the PS and these computations are then utilized at the PS to solve the optimization problem. One of the most popular centralized strategies, particularly for training deep neural network (DNN) architectures, is  {\em parallel synchronous stochastic gradient descent (PSSGD)}.
\begin{algorithm}[t]
\caption{PSSGD}\label{CSGD}
\begin{algorithmic}[1]
\For{$t=1,2,\ldots$}
    \For{$i\in \mathcal{S}$} in parallel
       \State Pull recent model $\boldsymbol{\theta}_{t}$ from PS
       \State Compute and send local SGD: $\mathbf{g}_{i,t}=\nabla_{\boldsymbol{\theta}}F(\boldsymbol{\theta}_{t},\zeta_{i,t})$
     \EndFor
    \State{\textbf{Model update}:} 
   \State $\boldsymbol{G}_{t}=\frac{1}{\vert \mathcal{S} \vert} \sum_{i\in\mathcal{S}} \mathbf{g}_{i,t}$
    \State $\boldsymbol{\theta}_{t+1}=\boldsymbol{\theta}_{t}-\eta_{t}\cdot \boldsymbol{G}_{t}$
\EndFor
\end{algorithmic}
\end{algorithm}

We recall that the objective in (\ref{DSO}) is written as the sum of $N$ functions; accordingly, the PSSGD works as follows: at the beginning of iteration $t$ each device pulls the current global model $\boldsymbol{\theta}_{t}$ from the PS, and computes the local gradient estimate to minimize its own loss function $f_{i}(\boldsymbol{\theta}_{t})$, i.e.,
\begin{equation} 
\mathbf{g}_{i,t}=\nabla_{\boldsymbol{\theta}}F(\boldsymbol{\theta}_{t},\zeta_{i,t}), \label{SGDAvg}
\end{equation}
where $\zeta_{i,t}$ is the data randomly sampled from the local dataset $\mathcal{D}_{i}$ in the $t$-th  iteration. We note here that the devices can also use mini-batches with multiple samples rather than a single data sample to compute the local gradient estimate. Once each device completes the computation of the local gradient estimate, the results are sent to the PS to be aggregated, and used to update the model parameter $\boldsymbol{\theta}$ as follows:
\begin{equation}
\boldsymbol{\theta}_{t+1}=\boldsymbol{\theta}_{t}-\eta_{t} \cdot \frac{1}{\vert \mathcal{S} \vert} \sum_{i\in\mathcal{S}} \mathbf{g}_{i,t},
\end{equation}
where $\eta_t$ is the learning rate at iteration $t$. 

The PSSGD algorithm is presented in Algorithm \ref{CSGD}. Here we note that, the algorithm is referred to as synchronous since the PS waits for all the devices to send their local gradient estimates before the model update, and at each iteration $t$, all the devices use the same global model to compute their local gradient estimates. This synchronisation may induce delays when the computation speed of some workers is slower than the others; hence, asynchronous variations of the centralized learning strategy have also been studied in the literature \cite{ASGD1,ASGD2,ASGD3}.

\subsection{Decentralized Learning}
Centralized learning is based on the star topology; that is, a central/ edge server, which can be an access point or a base station in the wireless context, orchestrates the devices participating in the collaborative training process. While this may be preferable in the presence of a base station that can communicate directly with all the devices in the coverage area, it may increase the burden on the cell-edge devices, and the multiple access channel from the devices to a single central server may become a bottleneck and limit the number of devices that can participate in the learning process at each iteration. Moreover, the star topology leads to a single point of failure, and increase the privacy and security risks. The alternative is fully decentralized learning, where the devices directly communicate with each other in a device-to-device fashion. Decentralized learning has been extensively studied in the literature \cite{decent3,decent5,decent6,decent7,decent9}. 

The main alteration in the decentralized scenario takes place in the consensus step, where the devices combine their local models with the neighboring ones according to a certain rule that is driven by the network topology. Therefore, compared to the star-topology used in centralized learning, the model at different devices are no longer fully synchronized after each global averaging step, introducing additional noise in the framework.  Formally speaking, each device $i$ seeks a consensus with the neighboring devices according to the following combining strategy:
\begin{equation}
\widetilde{\boldsymbol{\theta}}_{i,t}=\sum_{j\in\mathcal{S}} \mathbf{W}_{ij} \boldsymbol{\theta}_{j,t},
\end{equation}
where $\boldsymbol{\theta}_{j,t}$ denotes the model at device $j$ at iteration $t$, and $\mathbf{W}$ is a weight matrix. We set $\mathbf{W}_{ij}=0$ if there is no connection between device $i$ and device $j$. Under certain assumptions on $\mathbf{W}$, such as being {\em doubly stochastic}, convergence of the decentralized framework has been shown and its convergence speed is driven by the second largest {\em eigenvalue} of $\mathbf{W}$. The decentralized learning algorithm is presented in Algorithm \ref{alg:decent}.

		\begin{algorithm}[t]
		\caption{Decentralized learning}\label{alg:decent}
		\begin{algorithmic}
			\For{$t=1,\ldots,T$}
			\State{For  $i\in \mathcal{S}$} in parallel
			$\mathbf{g}_{i,t}=\nabla_{\boldsymbol{\theta}}\mathcal{L}_{i}(\boldsymbol{\theta}_{i,t},\zeta_{i,t})$
			\State{Consensus step: $\widetilde{\boldsymbol{\theta}}_{i,t}=\sum_{j\in\mathcal{S}} \mathbf{W}_{ij} \boldsymbol{\theta}_{j,t}$}
			\State{{Update local model}: $\boldsymbol{\theta}_{i,t}=\widetilde{\boldsymbol{\theta}}_{i,t}+\eta(t)\mathbf{g}_{i,t}$}
			\EndFor
		\end{algorithmic}
	\end{algorithm}
	
The weight matrix $\mathbf{W}$ can be constructed based on the network topology, which can be represented by a graph $\mathcal{G}$ in several ways. A common approach is to use the {\em graph Laplacian}, such that $\mathbf{W}$ is written as
\begin{equation}\label{weight}
	\mathbf{W}=\mathbf{I}-\frac{1}{d_{\max}+1}(\mathbf{D}-\mathbf{A}),
\end{equation}
where $\mathbf{A}$ is the adjacency matrix of the graph $\mathcal{G}$ representing the connectivity between the devices, $\mathbf{D}$ is a diagonal matrix whose entry in position $D_{ii}$ is the degree of the node corresponding to device $d_i$ in $\mathcal{G}$, and $d_{\max}$ is the maximum node degree. One can easily check that weight matrix designed according to (\ref{weight}) will be symmetric and doubly stochastic. For further details on the network topology and the convergence behavior readers can refer to \cite{decent.fund}, and to \cite{ozfatura2020decentralized} for the implementation of decentralized learning over a wireless network.

\section{Communication Efficient Distributed ML}

One of the main challenges in distributed learning is the communication bottleneck due to the large size of the trained models. Numerous communication efficient learning strategies have been proposed in the ML literature to reduce the number of bits exchanged between the devices and the PS per global iteration. We classify these approaches into three main groups; namely {\em sparsification}, {\em quantization}, and {\em local updates}, and present each of these approaches in detail in this section. We would like to highlight that these strategies are independent of the communication medium and the communication protocol employed to exchange model updates or gradient estimates between the devices and the PS, as they mainly focus on reducing the size of the messages exchanged. Therefore,  these techniques can be incorporated into the resource allocation and device selection policies that will be presented in Section \ref{s:Schedule_RA} when optimizing learning algorithms over wireless networks.




\subsection{Sparsification}

The objective of sparse SGD is to transform the $d$-dimensional gradient estimate $\mathbf{g}$ in (\ref{gradient_est}) to its sparse representation $\tilde{\mathbf{g}}$, where the non-zero elements of $\tilde{\mathbf{g}}$ are equal to the corresponding elements of $\mathbf{g}$. Sparsification can be considered as applying a $d$-dimensional mask vector $\mathbf{M}\in \left\{0,1\right\}^d$ on $\mathbf{g}$; that is,
\begin{equation}\label{sparsify}
\tilde{\mathbf{g}} = \mathbf{M}\otimes \mathbf{g},
\end{equation}
where $\otimes$ denotes element-wise multiplication. We denote the \textit{sparsification level} of this mask by
\begin{equation}
\phi \triangleq  \frac{\vert\vert\mathbf{M}\vert\vert_{1}}{d} << 1~.
\end{equation}
It has been shown that during the training of complex DNN architectures, such as ResNet \cite{NN.DRL} or VGG \cite{NN.VGG}, use of sparse SGD with  $\phi\in[0.01,0.001]$  provides a significant reduction in the communication load with minimal or no loss in the generalization performance \cite{SGD.sparse0,SGD.sparse1, SGD.sparse2, SGD.sparse3, SGD.sparse4, SGD.sparse5, SGD.sparse6, SGD.sparse7, SGD.sparse8}. Below, we will briefly overview some of the common sparsification strategies used in practice.\\

\subsubsection{Random sparsification}

In random sparsification, introduced in \cite{SGD.sparse1}, each element of vector $\mathbf{g}$ is set to zero independently with a prescribed probability. The $i$-th element of the sparsified vector $\mathbf{\tilde{g}}^{\text{rand}}$, for $i= 1, \ldots, d$, is obtained by
\[
    \mathbf{\tilde{g}}^{\text{rand}}_i =
\begin{cases}
    \frac{\mathbf{g}_{i}}{\mathbf{p}_{i}}, & \text{with probability } \mathbf{p}_{i}, \\
     0, & \text{with probability }1-\mathbf{p}_{i}.
\end{cases}
\]
This is an unbiased operator since 
\begin{equation}
\mathbb{E}[\mathbf{\tilde{g}}^{\text{rand}}_i]=\mathbf{p}_{i}\times\frac{\mathbf{g}_{i}}{\mathbf{p}_{i}}=\mathbf{g}_{i},
\end{equation}
and its variance is bounded by
\begin{equation}
\mathbb{E}\left[\sum^{d}_{i=1}\left(\mathbf{\tilde{g}}^{\text{rand}}_i\right)^2 \right] = \sum^{d}_{i=1}\frac{\mathbf{g}^{2}_{i}}{\mathbf{p}_{i}}.
\end{equation}
The key design issue behind random gradient sparsification is to minimize the number of non-zero values while controlling the variance; equivalently, solving the following optimization problem:
\begin{small}
\begin{align}
  \text{\bf P1:} & \;\;\; \min_{p_1, \ldots, p_d}
   \begin{aligned}[t]
       \sum^{d}_{i=1}\mathbf{p}_{i}  \notag
   \end{aligned} \\
  \text{subject to: }  \sum^{d}_{i=1}\frac{\mathbf{g}^{2}_{i}}{\mathbf{p}_{i}} &\leq (1+\epsilon)\sum^{d}_{i=1}\mathbf{g}^{2}_{i},
\end{align}
\end{small}
for some given $\epsilon > 0$. The solution of this optimization problem is given by \cite{SGD.sparse1}
\begin{equation}
\mathbf{p}_{i}=\min(\lambda \cdot \mathbf{g}_{i},1),
\end{equation}
where $\lambda$ is a parameter chosen based on $\epsilon$ and $\mathbf{g}$. 

\subsubsection{Synchronous sparse parameter averaging}

In this strategy, at each iteration, only a subset of the parameters are averaged to reduce the communication load. We call it as synchronous, since all the devices use an identical sparsification mask $\mathbf{M}_{t}$ at iteration $t$. 
Sparse parameter averaging strategy is executed in two steps \cite{SGD.sparse2}: In the first step, device $i$ updates its model based on the local gradient estimate
\begin{equation}
\boldsymbol{\theta}_{i,t+1/2}=\boldsymbol{\theta}_{i,t}-\eta_{t} \cdot \nabla_{\boldsymbol{\theta}}F(\boldsymbol{\theta}_{i,t},\zeta_{i,t}).
\end{equation}
Then, every device masks its local model with $\mathbf{M}_{t}$, and shares the masked parameters with the PS for aggregation. Accordingly, $\boldsymbol{\theta}_{i,t+1}$ evolves over the iterations in the following way:
\begin{equation}
\boldsymbol{\theta}_{i,t+1}=\frac{1}{N}\sum^{N}_{n=1}\boldsymbol{\theta}_{n,t+1/2}\otimes \mathbf{M}_{t}. 
\end{equation}
The convergence of the synchronous sparse parameter averaging strategy can be shown under the constraint that the sampling period of each parameter should not exceed some predefined value $\tau_{max}$; that is,
\begin{equation}
\vert\vert\otimes^{k}_{t=j}(1-\mathbf{M}_{t})\vert\vert_{1}=0,
\end{equation}
for any $k,j>0:k-j\geq\tau_{max}$.

\subsubsection{Top-$K$, Rand-$K$, and $R$-top-$K$ sparsification}

Top-$K$ sparsification is one of the most commonly used strategies for gradient sparsification. In the top-$K$ sparsification strategy each device constructs a sparsification mask $\mathbf{M}_{i,t}$ by identifying the indices of the largest $K$ values in $\vert\mathbf{g}_{i,t} \vert$ \cite{SGD.sparse3,SGD.local4,SGD.sparse5}. We use $S_{top}(\cdot,K)$ do denote this operation, i.e.,
\begin{equation}
\mathbf{M}_{i,t}=S_{top}(\vert\mathbf{g}_{i,t} \vert,K).
\end{equation}
The further details on the convergence of top-$K$ sparsification can be found in \cite{SGD.local4,SGD.sparse5}. 

The rand-$K$ sparsification strategy \cite{SGD.sparse5} selects the sparsification mask $\mathbf{M}_{i,t}$ randomly from the set of masks with sparsification level $K$:
\begin{equation}
\mathcal{M}=\left\{\mathbf{M}:\left\{0,1\right\}^d,\vert\vert\mathbf{M}\vert\vert_{1}=K\right\}.
\end{equation}
We note that, unlike the random sparsification strategy in \cite{SGD.sparse1}, rand-$K$ and top-$K$ do not result in an unbiased compression strategy. In the case of rand-$K$, scaling $\mathbf{M}_{i,t}$ with $d/K$ can be used to obtain unbiased compressed gradient estimates; however, such correction also scales the variance of the compression scheme, thus may not be desirable in practice \cite{SGD.sparse5}. In general, top-$K$ sparsification strategy has been shown to perform better compared to rand-$K$ in practice in terms of the test accuracy and convergence speed. However, top-$K$ sparsification comes with the additional complexity due to the sorting of the elements of $\vert\mathbf{g}_{i,t} \vert$. It also requires increased communication load since the indices of the non-zero elements of each mask vector $\mathbf{M}_{i,t}$ should also be transmitted. This may not be a requirement for rand-$K$ if a pseudo-random generator with a common seed is used so that all the devices can generate the same mask. 

We also note that, when employed for distributed training of DNN architectures, these sparse communication strategies can be applied to each layer of the network separately, since it is observed that different layers have different tolerance to sparsification of their weights. We refer the readers to \cite{SGD.sparse0} for further information on gradient sparsification. Finally, a hybrid approach, called $R$-top-$K$, is considered in \cite{SGD.sparse6}. In this scheme, each device first identifies the largest $R$ values in $\vert\mathbf{g}_{i,t}\vert$, and then selects $K$ of them randomly. This hybrid approach results in better compression as well as less bias.

We also want to remark that the use of top-K sparsification locally, that is having a separate mask for each worker, has the drawback that the sparsification is achieved only in the uplink direction, whereas the global model update from the PS to the devices will not be sparse. Moreover, the mismatch between the sparsity patterns across the workers limits the use of more efficient communication protocols, such as {\tt all-reduce}. This problem is addressed in \cite{SGD.sparse10}, where a majority voting based strategy is proposed to seek a consensus among the workers to form a common sparsification mask.

\begin{algorithm}[t]\label{SSGDEF}
\caption{Sparse SGD with error accumulation}
\begin{algorithmic}[1]
    \For{$t=1,\ldots,T$}
        \State \underline{\textbf{ Device side:}}
        \For{$n=1,\ldots,N$} in parallel
        \State Receive $\widetilde{\mathbf{G}}_{t-1}$ from PS
        \State \textbf{Update model:} $\boldsymbol{\theta}_{t} = \boldsymbol{\theta}_{t-1} + \eta_{t}\widetilde{\mathbf{G}}_{t-1}$
        \State \textbf{Compute SGD:} $\mathbf{g}_{n,t}=\nabla_{\boldsymbol{\theta}}F(\boldsymbol{\theta}_{t},\zeta_{n,t})$
        \State \textbf{Sparsification with error accumulation:}
        \State{$\widetilde{\mathbf{g}}_{n,t}=\mathbf{M}_{n,t}\otimes(\mathbf{g}_{n,t}+\mathbf{e}_{n,t}$})
        \State \textbf{Update the error:}
 \State{$\mathbf{e}_{n,t+1}=(1-\mathbf{M}_{n,t})\otimes(\mathbf{g}_{n,t}+\mathbf{e}_{n,t}$})
        \State Send $\widetilde{\mathbf{g}}_{n,t}$ to PS
        \EndFor
    \State\underline{\textbf{Sparse communication PS side:}}
    \State \textbf{Aggregate local sparse gradient:}
    \State $\widetilde{\mathbf{G}}_{t}=\sum_{n\in[N]}\widetilde{\mathbf{g}}_{n,t}$
    \State send $\widetilde{\mathbf{G}}_{t}$ to all devices
    \State \textbf{Aggregate local sparse gradients with error accumulation:}
    \State $\widetilde{\mathbf{G}}_{t}=\mathbf{M}_{n,t}\otimes(\sum_{n\in[N]}\widetilde{\mathbf{g}}_{n,t}+\mathbf{e}_{t})$
    \State Send $\widetilde{\mathbf{G}}_{t}$ to all devices
    \State \textbf{Update error:}
    \State $\mathbf{e}_{t+1}=(1-\mathbf{M}_{t})\otimes(\widetilde{\mathbf{G}}_{n,t}+\mathbf{e}_{n,t})$
    \EndFor
\end{algorithmic}
\end{algorithm}

\subsubsection{Error accumulation}

The idea behind the error accumulation is to compensate in each iteration for the compression error that has been introduced in the previous iteration by simply adding the previous compression error to the current gradient estimate. It has been shown that \cite{SGD.sparse3,SGD.local4,SGD.sparse5} error accumulation improves the convergence performance. When error accumulation is employed, the sparsification step at the devices is rewritten as
\begin{equation}
\widetilde{\mathbf{g}}_{n,t}=\mathbf{M}_{n,t}\otimes(\mathbf{g}_{n,t}+\mathbf{e}_{n,t}),
\end{equation}
where $\mathbf{e}_{n,t}$ is the accumulated error from the previous iteration. Following gradient sparsification, the compression error is updated as follows:
\begin{equation}
\mathbf{e}_{n,t+1}=(1-\mathbf{M}_{n,t})\otimes(\mathbf{g}_{n,t}+\mathbf{e}_{n,t}).
\end{equation}

The overall sparse SGD framework with error accumulation is illustrated in Algorithm \ref{SSGDEF}. We remark that gradient sparsification can be employed both in the device-to-PS (uplink) and PS-to-device (downlink) directions to improve the communication efficiency further. In Algorithm \ref{SSGDEF}, sparsification at the PS side is applied in lines 16-20. A further analysis on this double sparsification strategy can be found in \cite{SGD.sparse9}. In general, unbiased compression schemes are preferred for distributed learning as they lend themselves to theoretical convergence analysis; however, unbiased sparsification methods typically suffer from large variation, making them less desirable for practical applications \cite{SGD.sparse5}.

The key benefit of the error accumulation mechanism is to make a biased sparsification method an efficient compression strategy for distributed learning, backed with a theoretical convergence analysis. It is shown in \cite{SGD.sparse5} that a gradient sparsification scheme with error accumulation converges if the it is in the form of a {\em $k$-contraction operator}, which is defined below.
\begin{definition}
 For a parameter $0<k\leq d$, a \textit{$k$-contraction operator} is a (possibly randomized) operator $\mathrm{comp}: \mathbb{R}^{d}\mapsto \mathbb{R}^{d}$ that satisfies the contraction property
\begin{equation}
\mathbb{E}\left[\vert\vert \mathbf{x}- \mathrm{comp}(\mathbf{x})\vert\vert^{2} \right] \leq \left(1-\frac{k}{d}\right) \vert\vert \mathbf{x} \vert\vert^{2}, ~~~ \forall \mathbf{x}\in\mathbb{R}^{d}.
\end{equation}
\end{definition}
In \cite{SGD.sparse5}, it is also shown that both the rand-$K$ and top-$K$ sparsification schemes are indeed $k$-contraction operators. The notion of compressor operation is further generalized in \cite{SGD.feedback}.

\begin{algorithm}
\caption{Sparse position decoding}\label{alg:encode}
\begin{algorithmic}[1]
\State{Initialize: $pointer=0$}
\While{$pointer<length(\mathbf{v}_{loc})$}
\If{$\mathbf{v}_{loc}(pointer)=0$}
\State{$blockindex=blockindex+1$}
\State{$pointer=pointer+1$}
\Else
\State{Recover $IntraBlockPosition$:}
\State{Read the next $\log_2(1/\phi)$ bits}
\State{Recover the location of a non-zero value:}
\State{$(1/\phi)\cdot blockindex+IntraBlockPosition$}
\State{$pointer=pointer + \log_2 (1/\phi) +1$}
\EndIf
\EndWhile
\end{algorithmic}
\end{algorithm}

\subsubsection{Sparse representation}

When a sparsity mask is employed to reduce the communication load in distributed learning, the mask also needs to be communicated to the PS together with the non-zero entries. In general, $\log_2 d$ bits are needed to convey the position of each non-zero element of a sparse masking vector. However, this number can be reduced by employing the following coding scheme. Let $\mathbf{v}$ be a sparse vector with sparsification level $\phi$. Assume that $\mathbf{v}$ is divided into equal-length blocks of size $1/\phi$. Position of a non-zero value within a particular block can be represented by only $\log_2 (1/\phi)$ bits. In order to identify each block we use an additional bit in the following way: we use $0$ to skip to the next block, and $1$ to indicate that the next $\log_2 (1/\phi)$ bits represent the location within the current block. Hence, on average $\log_2 (1/\phi) + 1$ bits are needed to represent the location of each non-zero value, and a total of $\phi d$ bits to determine the block indices.

In the decoding part, given the encoded binary vector $\mathbf{v}_{loc}$ for the position of the non-zero values in $\mathbf{v}$, the PS starts reading the bits from the first index and checks whether it is $0$ or $1$. If it is $1$, then the next $\log_2 (1/\phi)$ bits are used to recover the position of the non-zero value in the current block. If the index is $0$, then the PS increases the {\em blockindex}, which tracks the current block, by one, and moves on to the next index. The overall decoding procedure is illustrated in Algorithm \ref{alg:encode}.

For example, consider a sparse vector of size $d=24$ and $\phi=1/8$; that is, there are only 3 non-zero values, and let those values be at indices $1, 5, 17$. Since $\phi=1/8$, the vector is divided into 3 blocks, each of size $8$, and the position of non-zero values within each block can be represented by 3 bits, i.e., $000,100,000$; hence, overall the sparse representation can be written as a bit stream of
\begin{equation}
\mathbf{1}000\mathbf{1}100\underline{0}\underline{0}\mathbf{1}000\underline{0},
\end{equation}
where the bits in normal font represent the position within the block, bold bits indicate that the following three bits refer to the position within the current block, and the underlined bits represent the end of a block. We note that other methods such {\em Elias coding} \cite{SGD.qs1} and {\em Golomb coding} \cite{FL.sparse1} can also be used for an efficient representation of the sparse mask vector.

\subsection{Quantization}
Even after applying a sparsifying transform as in (\ref{sparsify}), we inherently assume that the exact values of the local updates, $\tilde{\mathbf{g}}$, are communicated from the devices to the PS. In practice, the communication links from the devices to the PS are finite capacity, and hence, these real-valued vectors must be quantized. In the standard setting, floating point precision with 32 bits is assumed to represent each real value, thus 32 bits are required to represent each element of the local gradient estimate. Quantization techniques aim to represent each element of the gradient estimates with as few bits as possible to reduce the communication load \cite{SGD.qs1,SGD.qs2,SGD.qs3,SGD.q0, SGD.q1, SGD.q2, SGD.q3, SGD.q4, SGD.q5, SGD.q6}. In the most extreme case, only the sign of each element is sent, i.e., using only a single bit per dimension, to achieve up to $\times32$ reduction in the communication load \cite{SGD.q0, SGD.q4, SGD.q5, SGD.q6}.

\subsubsection{Stochastic Uniform Quantization}

\indent In stochastic uniform quantization strategy \cite{SGD.qs1, SGD.qs2}, denoted by $\mathcal{Q}_{s}(\cdot)$, we uniformly allocate representation points as in uniform quantization, but instead of mapping each number to the nearest representation point, we probabilistically assign them to one of the two nearest points. We first normalize and take the absolute values of the elements of the given vector $\mathbf{u}$: $\mathbf{u}\in\mathbb{R}^{d}$:
\begin{equation}
\tilde{\mathbf{u}}=\frac{\vert\mathbf{u}\vert}{\vert\vert\mathbf{u}\vert\vert}.
\end{equation}
Then, consider dividing the interval $[0,1]$ into $L$ equal length sub-intervals $I_{1},\ldots,I_{L}$, where $I_{l}=[I^{lower}_{l},I^{upper}_{l})$. In the last step, each normalized value  $\tilde{\mathbf{u}}_{i}$ is randomly mapped to one of the boundary points of the corresponding interval according to $q_{rand}(\cdot)$ and scaled by the sign of $\mathbf{u}_{i}$. Let $\tilde{\mathbf{u}}_{i}\in I_{l}$; we have
\[
    q_{rand}(\tilde{\mathbf{u}}_{i})=
\begin{cases}
    I^{lower}_{l}, & \text{with probability } (I^{upper}_{l}-\tilde{\mathbf{u}}_{i})L\\
     I^{upper}_{l}, & \text{with probability }(\tilde{\mathbf{u}}_{i}-I^{lower}_{l})L.
\end{cases}
\]
Hence, overall, the quantized representation of $\mathbf{u}$ can be written as
\begin{equation}
\mathcal{Q}_{s}(\mathbf{u})_{i}= \mathrm{sign}(\mathbf{u}_i) q_{rand}(\tilde{\mathbf{u}}_{i})\vert\vert\mathbf{u}\vert\vert 
\end{equation}
In \cite{SGD.qs3}, stochastic quantization framework is combined with the error accumulation mechanism.

\subsubsection{Ternary gradient quantization}

Similarly to stochastic gradient quantization, ternary gradient quantization is another unbiased compression strategy, which can be considered as a mixture of random sparsification and stochastic quantization \cite{SGD.q5}. It is called ternary since the quantized values can take only three values, i.e.,
\begin{equation}
Q_{tern}(\mathbf{g})_{i}\in\left\{-1,0,+1\right\}.
\end{equation}
It can be represented as follows:
\begin{equation}
Q_{tern}(\mathbf{g})=g_{max} \cdot \mathrm{sign}(\mathbf{g})\otimes\mathbf{b}
\end{equation}
where
\begin{equation}
g_{max}=\vert\vert\mathbf{g}\vert\vert_{\infty},
\end{equation}
and $\mathbf{b}$ is a random binary vector of independent Bernoulli variables, where
\[
    \mathbf{b}_{i}=
\begin{cases}
    1, & \text{with probability } \frac{\vert\mathbf{g}_{i}\vert}{g_{max}}, \\
     0, & \text{with probability }1-\frac{\vert\mathbf{g}_{i}\vert}{g_{max}}.
\end{cases}
\]

\begin{algorithm}
\caption{SignSGD}
\label{SignSGD}
\begin{algorithmic}[1]
    \For{$t=1,\ldots,T$}
        \State \underline{\textbf{ Device side:}}
        \For{$n=1,\ldots,N$} in parallel
        \State \textbf{Update model:} $\boldsymbol{\theta}_{t} = \boldsymbol{\theta}_{t-1} + \eta \times \mathrm{sign}\left(\sum_{n\in[N]} \mathrm{sign}(\mathbf{g}_{n,t})\right)$
        \State \textbf{Compute SGD:} $\mathbf{g}_{n,t}=\nabla_{\boldsymbol{\theta}}F(\boldsymbol{\theta}_{t},\zeta_{n,t})$
        \State Send $\mathrm{sign}(\mathbf{g}_{n,t})$ to PS
        \EndFor
    \State\underline{\textbf{Sparse communication PS side:}}
    \State Send $\mathrm{sign} \left(\sum_{n\in[N]} \mathrm{sign}(\mathbf{g}_{n,t})\right)$ to devices
    \EndFor
\end{algorithmic}
\end{algorithm}

\subsubsection{SignSGD and 1-bit quantization}

1-bit quantization strategy \cite{SGD.q0,SGD.q1} is one of the earliest schemes that have been employed in distributed learning to reduce the communication load, where each device quantizes its gradient using a single threshold value so that only a single bit is required to represent each element.

A similar approach that uses only one bit for the gradient values is {\em SignSGD} \cite{SGD.q2}, where each device only sends the sign of the gradient values. Equivalently, in SignSGD the distributed learning framework works as a majority vote since the sum of the signs indicates the result of a majority vote for the gradient direction. The SignSGD algorithm is presented in Algorithm \ref{SignSGD}. We note that, thanks to the reduced dimensionality, SignSGD strategy also provides certain robustness against adversarial attacks \cite{SGD.q3}.
\subsubsection{Scaled sign operator}
Another quantization scheme is the scaled sign operator, which has a similar structure to ternary quantization, although it is not an unbiased compression scheme. The scaled sign operator sends the sign of the gradient values by scaling them with the mean value of the gradient, i.e.,
\begin{equation}\label{scale_sign}
\mathcal{Q}(\mathbf{\mathbf{g}})= \frac{\vert\vert \mathbf{g}\vert\vert_{1}}{d}  \mathrm{sign}(\mathbf{g}),
\end{equation}
where $d$ is the dimension of vector $\mathbf{g}$. Although the scaled sign operator is biased, it is a $\delta$-approximate compressor, that is, it satisfies the following inequality:
\begin{equation}
\vert\vert \mathcal{Q}(\mathbf{x})- \mathbf{x}\vert\vert^{2}_{2}\leq 1-\delta \cdot \vert\vert \mathbf{x}\vert\vert^{2}_{2}.
\end{equation}
In \cite{SGD.q4}, it is shown that when a quantization scheme is a $\delta$-approximate compressor, then the quantized SGD framework converges when the quantization scheme is employed together with error accumulation. 

Block-wise implementation of the scaled sign operator is considered in \cite{SGD.q5}, where the gradient vector $\mathbf{g}$, or the momentum term if momentum SGD is used as an optimizer, is divided into $L$ disjoint smaller blocks $\left\{\mathbf{g}_{1},\ldots, \mathbf{g}_{L} \right\}$, and the scaled sign operator is applied to each block separately. This way, variations on the mean gradient values of different layers can be incorporated to the compression procedure to further reduce the quantization error. We also note that in \cite{SGD.q5} quantization is employed both in uplink and downlink direction. An efficient compression scheme for transmission in the PS-to-device direction is presented in \cite{amiri2020convergence}.

\begin{algorithm}
\caption{Compressed local SGD with error accumulation }\label{alg:SSGD_MV}
\begin{algorithmic}[1]
    \For{$t=1,\ldots,T$}
        \State \underline{\textbf{ Device side:}}
        \For{$n=1,\ldots,N$} in parallel
        \State Receive $\widetilde{\mathbf{\Delta}}_{t-1}$ from PS
        \State \textbf{Update model:} $\boldsymbol{\theta}_{n,t} = \boldsymbol{\theta}_{n,t-1} + \widetilde{\boldsymbol{\Delta}}_{t-1}$
        \State Initialize $\boldsymbol{\theta}^{0}_{n,t}=\boldsymbol{\theta}_{n,t}$
        \State \textbf{Perform local SGD for $H$ iteartions:} 
        \State $\mathbf{\Delta}_{n,t}=\boldsymbol{\theta}^{H}_{n,t}-\boldsymbol{\theta}^{0}_{n,t}$
        \State{$\widetilde{\mathbf{\Delta}}_{n,t}=C(\mathbf{\Delta}_{n,t} +\mathbf{e}_{n,t}$})
        \State Send $\widetilde{\mathbf{\Delta}}_{n,t}$ to PS
        \State $\mathbf{e}_{n,t}= \mathbf{\Delta}_{n,t}-\widetilde{\mathbf{\Delta}}_{n,t}$
        \EndFor
    \State\underline{\textbf{Sparse communication PS side:}}
    \State $\widetilde{\mathbf{\Delta}}_{t}=\sum_{n\in[N]}\widetilde{\boldsymbol{\Delta}}_{n,t}$
    \State Send $\widetilde{\mathbf{\Delta}}_{t}$ to all devices
        \State $\widetilde{\mathbf{\Delta}}_{t}=C(\sum_{n\in[N]}\widetilde{\boldsymbol{\Delta}}_{n,t}+\mathbf{e}_{t})$
    \State Send $ \widetilde{\mathbf{\Delta}}_{t}$ to all devices
    \State $\mathbf{e}_{t+1}=\sum_{n\in[N]}\widetilde{\boldsymbol{\Delta}}_{n,t}+\mathbf{e}_{t}-\widetilde{\mathbf{\Delta}}_{t}$
    \EndFor
\end{algorithmic}
\end{algorithm}

\subsection{Local iteration}
Local SGD, also known as \textit{$H$-step averaging}, aims to reduce the frequency of global model updates at the PS, and hence, the communication of local model updates from the devices to the PS \cite{SGD.local1, SGD.local2, SGD.local3, SGD.local4, SGD.local5, SGD.local6, SGD.local7}. In order to reduce the frequency of global aggregation, devices update their models locally for $H$ consecutive SGD iterations based on their local gradient estimates, before sending their local model updates to the PS. The global model evolves over communication rounds as follows:
\begin{equation} \label{model_update_mult}
\boldsymbol{\theta}_{t+1}= \boldsymbol{\theta}_{t}+\frac{1}{N}\sum^{N}_{n=1}\underbrace{\sum^{H}_{h=1}-\eta_{t} \cdot \mathbf{g}^{h}_{n,t}}_{\Delta\boldsymbol{\theta}_{n,t}}, 
\end{equation}
where
\begin{equation}
\mathbf{g}^{h}_{n,t}=\nabla_{\boldsymbol{\theta}}F(\boldsymbol{\theta}^{h}_{n,t},\zeta_{n,h}),
\end{equation}
and
\begin{equation}
\boldsymbol{\theta}^{h}_{n,t}=\boldsymbol{\theta}^{h-1}_{n,t}-\eta_{t} \cdot \mathbf{g}^{h}_{n,t},
\end{equation}
where we set $\boldsymbol{\theta}^{0}_{n,t} = \boldsymbol{\theta}_{t}$.

In this section, we have presented three groups of strategies to reduce the communication load in distributed learning. We would like to highlight that these strategies can be combined to reduce the communication load further \cite{SGD.comb1,FL.sparse1}. In Algorithm \ref{alg:SSGD_MV}, a general communication efficient distributed learning framework is presented where $C$ can be one of the compression operators previously explained or a combination of these schemes. We note that the lines 15-17 in Algorithm \ref{alg:SSGD_MV} are used instead of lines 13 and 14 when the compression strategy is also employed in the downlink direction.

\begin{algorithm}[t]
\caption{Federated Averaging (FedAVG)(FL)}\label{alg:dist}
\begin{algorithmic}[1]
\For{$t=1,2,\ldots$}
    \State{Choose a subset of devices $\mathcal{S}_{t}\subseteq \mathcal{S}$ 
    } 
    \For{$n\in \mathcal{S}_{t}$} in parallel
        \State Pull $\boldsymbol{\theta}_{t-1}$ from PS: $\boldsymbol{\theta}^{0}_{n,t}=\boldsymbol{\theta}_{t-1}$
        \For{$h=1,\ldots,H$}
            \State Compute SGD: $\mathbf{g}^{h}_{n,t}=\nabla_{\boldsymbol{\theta}}F(\boldsymbol{\theta}^{h-1}_{n,t},\zeta_{n,h})$
            \State Update model: $\boldsymbol{\theta}^{h}_{n,t}=\boldsymbol{\theta}^{h-1}_{n,t}-\eta_{t}\mathbf{g}^{h}_{n,t}$
       \EndFor
       \State Push $\boldsymbol{\theta}^{H}_{n,t}$
    \EndFor
    \State{\textbf{Federated Averaging}:} $\boldsymbol{\theta}_{t}=\frac{1}{\vert \mathcal{S}_{t} \vert} \sum_{i\in\mathcal{S}_{t}} \boldsymbol{\theta}^{H}_{n,t}$
\EndFor
\end{algorithmic}
\end{algorithm}

\subsection{Federated Learning (FL)}
FL is a centralized distributed learning strategy, which was introduced to solve the stochastic optimization problem given in (\ref{DSO}) in a distributed manner without sharing local datasets in order to offer a certain level of privacy to users. This is particularly relevant when sensitive data, such as personal medical records, are used to train a model, e.g., for digital healthcare /remote diagnosis applications \cite{FL.health1,FL.health2}.

From the  implementation perspective, FL framework differs from other centralized learning strategies since it aims to practice collaborative learning on a large scale. Therefor, for practical implementation FL requires some additional mechanisms to make it communication efficient. The first mechanism used in FL framework is the use of local SGD strategy to reduce the communication frequency between the PS and the devices. Second, to prevent overload at the PS, at each iteration $t$, only a subset of the active devices in the network, denoted by $\mathcal{S}_{t} \subseteq \mathcal{S}$, are selected to participate the learning process. The devices can be sampled randomly, or by some more advanced selection schemes as we will address later.

In a broad sense, FL employs distributed SGD, and each SGD iteration consists of three main steps, namely: {\em device sampling/selection}, {\em local computation}, and {\em consensus/aggregation}. Once the PS samples the devices $\mathcal{S}_{t}\subseteq\mathcal{S}$ for the $t$-th iteration, each device first pulls the latest global model $\boldsymbol{\theta}_{t-1}$ from the PS, and then locally minimizes its own loss function $f_{n}(\boldsymbol{\theta}_{t})$ with an SGD update, i.e.,
\begin{equation}\label{SGD}
\boldsymbol{\theta}_{n,t}^{h}=\boldsymbol{\theta}_{n,t}^{h-1}-\eta_{t} \cdot \mathbf{g}^{h}_{n,t}
\end{equation}
for $h=1, \ldots, H$, where we set $\boldsymbol{\theta}_{i,t}^0 = \boldsymbol{\theta}_{t-1}$, 
\begin{equation} 
\mathbf{g}^{h}_{n,t}=\nabla_{\boldsymbol{\theta}}F(\boldsymbol{\theta}_{n,t}^h,\zeta_{n,h}),
\end{equation}
and $\zeta_{n,h}$ is the data sampled from dataset $\mathcal{D}_{n}$ in the $h$-th local iteration.

In each iteration, each device carries out $H$ SGD steps before the aggregation at the PS. Once each device finishes its local updates, it sends the latest local model $\boldsymbol{\theta}_{n,t}^H$ to the PS for aggregation. The common approach for the aggregation/consensus in FL is to simply take the average of the participating devices' local models, i.e.,
\begin{equation}
\boldsymbol{\theta}_{t}=\frac{1}{\vert \mathcal{S}_{t} \vert} \sum_{i\in\mathcal{S}_{t}}\boldsymbol{\theta}_{i,t}^H.
\end{equation} 
However, we remark here that more advanced aggregation approaches have also been introduced recently in \cite{FedMA, FedED, FedDist}. The generic FL framework iterating over these three steps, called as {\em federated averaging (FedAVG)}, is provided in Algorithm \ref{alg:dist}. We note that the particular scenario with full participation  i.e., $\mathcal{S}_{t}=\mathcal{S}$, and $H=1$, is referred to as {\em federated SGD (FedSGD)}.

\subsubsection{Accelerated FL} 

There have been many efforts to accelerate FL convergence. A classical technique for accelerating SGD is the momentum method \cite{Polyak}, which accumulates a velocity vector in the directions of persistent reduction in the objective function across iterations. State of the art results for most ML tasks incorporate momentum. Momentum can also be incorporates into distributed SGD in various ways. \textit{Global momentum} can be used at the PS when updating the global model, or the devices can apply \textit{local momentum} during their local update steps, or a combination of the two can also be used \cite{SGD.local2}. 

We can treat the FL framework as a combination of two loops, where in the inner loop devices update their local models, while in the outer loop PS updates the global model. In principle, it is possible to accelerate the convergence of FL by using different optimizers in either of these loops. One such strategy is the SlowMo framework \cite{FL.acc1}, which is summarized in Algorithm  \ref{alg:SlowMo}. The core idea behind the SlowMo framework is to utilize the local model updates from the inner loop to compute a {\em pseudo gradient} at the PS, which is then used by the PS as the optimizer for the outer loop to update the global model. Formally speaking, each device scheduled for the $t$-th iteration of the FL algorithm performs $H$ local updates as before, and sends the accumulated model difference $\boldsymbol{\Delta}_{n,t} \triangleq \boldsymbol{\theta}^{H}_{i,t}- \boldsymbol{\theta}_{t-1}$ to the PS (line 10 of Algorithm \ref{alg:SlowMo}). PS utilizes the average of the model updates as the pseudo gradient and updates the momentum  of the outer loop accordingly (line 14 of Algorithm \ref{alg:SlowMo}), which is then used to update the model (line 16 of Algorithm \ref{alg:SlowMo}). We note that SlowMo uses {\em momentum SGD} at the PS, but this approach can also be extended to other optimizers \cite{FL.acc2}.

\begin{algorithm}[t]
\caption{SlowMo}\label{alg:SlowMo}
\begin{algorithmic}[1]
    \For{$t=1,\ldots,T$}
    \State \underline{\textbf{Local iteration:}}
    \State Sample participating devices $\mathcal{S}_{t}\subseteq\mathcal{S}$
     \For{$i\in \mathcal{S}_{t}$} in parallel
      \State $\boldsymbol{\theta}^{0}_{i,t}=\boldsymbol{\theta}_{t-1}$
        \For{$h=1,\ldots,H$} local update:
            \State Compute SGD: $\mathbf{g}^{h}_{i,t}=\nabla_{\boldsymbol{\theta}}F(\boldsymbol{\theta}^{h-1}_{i,t},\zeta_{i,h})$
            \State Update model: $\boldsymbol{\theta}^{h}_{i,t}=\boldsymbol{\theta}^{h-1}_{i,t}-\eta_{t}\mathbf{g}^{h}_{i,t}$
       \EndFor
    \EndFor
\State\underline{\textbf{Communication phase:}}
\For{$i \in\mathcal{S}_{t}$}
\State{Send to PS: $\boldsymbol{\Delta}_{i,t} \triangleq \boldsymbol{\theta}^{H}_{i,t} - \boldsymbol{\theta}_{t-1}$}
\EndFor
\State \underline{\textbf{Compute pseudo gradient:}}
\State $\bar{\mathbf{G}}_{t} = \frac{1}{\vert\mathcal{S}_{t}\vert}\frac{1}{\eta_{t}}\sum_{i\in\mathcal{S}_{t}} \boldsymbol{\Delta}_{i,t}$
\State \underline{\textbf{Compute pseudo momentum:}}
\State $\mathbf{m}_{t+1}=\beta\mathbf{m}_{t} + \bar{\mathbf{G}}_{t}$
\State\underline{\textbf{Model update:}}
\State $\boldsymbol{\theta}_{t+1}=\boldsymbol{\theta}_{t}-\alpha\eta_{t}\mathbf{m}_{t+1}$
\EndFor
\end{algorithmic}
\end{algorithm}

\section{Device Selection and Resource Allocation in Distributed Learning Over Wireless Networks}\label{s:Schedule_RA}

A fundamental design problem in centralized collaborative learning is to find the optimal {number of uplink devices} and the optimal {frequency} of global model updates, or equivalently, the number of local iterations in between consecutive global updates, to seek a balance between the accuracy of the model update and the communication cost in order to achieve the best convergence performance based on the wall clock time. Collaborative learning across wireless devices introduces new challenges as the communication bottleneck becomes even more stringent due to the limited bandwidth available for model updates, varying nature of the channel quality across users and iterations, and the energy-limited nature of most wireless devices. Hence, the optimal collaborative learning design across wireless networks must take into account the channel conditions of the devices, and optimize the wireless networking parameters together with the parameters of the collaborative learning algorithm \cite{FL.CS-RA3, FL.CS-RA4, FL.CS-RA5, FL.CS-RA6}.

The two most common device selection mechanisms employed in the FL framework are \textit{random scheduling}, where devices participating in each iteration are randomly sampled, and \textit{round-robin scheduling}, where the devices are divided into groups and the groups are scheduled according to a predefined order throughout the learning process. However, these scheduling mechanisms do not take into account the channel condition of each individual device, and since the global model update requires certain synchronization across selected devices at each iteration, devices with weaker channel conditions may become the bottleneck, resulting in significant delays.

To overcome the aforementioned issue, device selection mechanism in FL should be modified by taking into account the channel conditions of the devices. Let $L^{\text{comm}}_{i,t}$ and  $L^{\text{comp}}_{i,t}$ be the communication and computation latency of the $ith$ device at  iteration $t$, respectively. We have $L^{\text{comm}}_{i,t}=d/R_{i,t}$, where $d$ is the size of the local update in bits, and $R_{i,t}$ (bits per unit time) is the transmission rate of the $i$th device at iteration $t$. The transmission rate of device $i$, $R_{i,t}$, is a function of its transmission power $P_{i,t}$, channel state $h_{i,t}$, and the resource allocation policy $\boldsymbol{\Phi}_{t}$, which decides on the time/frequency resources allocated to each device at each iteration. From the latency perspective, collaborative learning in a wireless network setup can be formulated as a joint device selection ($\boldsymbol{\Pi}_{t}$) and power/resource allocation ($\mathbf{P}_{t},\boldsymbol{\Phi}_{t}$) problem, where the objective can be written, generically, in the following form:
\begin{equation}
\min_{\mathbf{P}_{t},\boldsymbol{\Pi}_{t},\boldsymbol{\Phi}_{t}\vert\mathbf{h}_{t}}\max_{i \in \boldsymbol{\Pi}_{t}} \underbrace{\frac{d}{R_{i,t}(\mathbf{P}_{t}, \mathbf{h}_{t}, \boldsymbol{\Phi}_{t})}}_{L^{\text{comm}}_{i,t}} + L^{\text{comp}}_{i,t},
\label{opt1}
\end{equation}
where $\boldsymbol{\Pi}_{t} \subseteq \mathcal{S}$ denotes the set of scheduled devices at iteration $t$. We note that the objective in ($\ref{opt1}$) depicts a general form, while in a practical scenario we may assume fixed transmission power, in which case the objective function reduces to joint device selection and resource allocation problem, and it can be further simplified to only a device selection problem by assuming a uniform, or a fixed resource allocation strategy. We can also impose a constraint on the number of scheduled devices at each round, i.e., $|\boldsymbol{\Pi}_{t}| = K$, $\forall t$, for some $0<K \leq \vert\mathcal{S} \vert$, or leave it as an optimization parameter. 



\begin{figure*}[h]
\begin{center}
\includegraphics[scale =0.4]{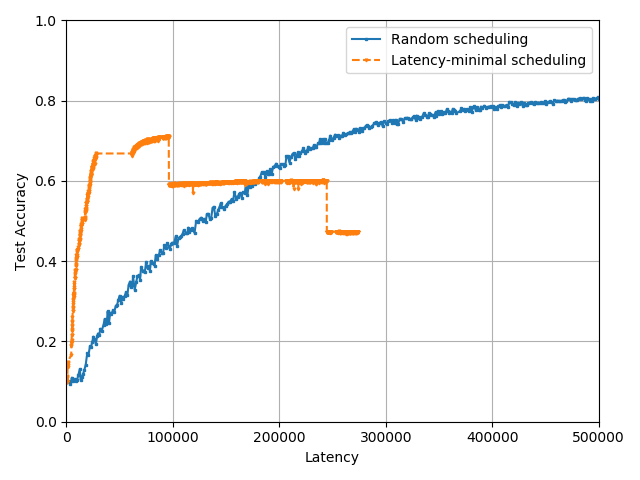}
\caption{Test accuracy with respect to overall latency (seconds) analyzed over 1000 communication rounds for collaborative learning of classification on the CIFAR-10 dataset using a CNN architecture. In latency-minimal scheduling, we schedule the users with the best channel states at each iteration. }
\label{randvsmrtp}
\end{center}
\end{figure*}

In the FL context, it is possible to establish a connection between the number of scheduled devices and the number of communication rounds to achieve a certain level of final accuracy. However, this assumes that the devices are scheduled independently and with identical probability, such that, although one device may be preferred over another at a certain iteration $t$, devices participate in the model updates equally in the long run in order not to introduce any bias in the overall process. However, in the wireless setting, scheduling statistics may not be identical for devices; for example, when the access point acting as the PS is not equally distant to the devices. Such heterogeneity requires further constraints on communication efficient device selection strategies for FL. To clarify this point, we present an example which highlights why purely channel-dependent scheduling strategies are not sufficient in the FL framework.\\

\indent In this example, we consider training a CNN architecture in a federated manner with the help of a single-antenna base station. We assume there are 100 single-antenna edge devices uniformly distributed around the base station within a radius of 500 meters, and at each iteration 20 devices participate in the global model update. In the simulation, we consider both large-scale path loss and small-scale Rayleigh fading with unit power, and consider transmission at the corresponding channel capacity assuming a random channel realization at each iteration. 
We assume both BS and devices transmits with a fixed power of 15dbm and 10 dbm, respectively, and we set the bandwidth to $B=2\times10^{7}$ and the noise power to $N_{0}=-204$ dbW/Hz. 

We compare random scheduling with channel-aware scheduling, which schedules the devices in order to minimize the latency as in (\ref{opt1}). We plot the test accuracy of the learned model throughout the learning process in Fig. \ref{randvsmrtp}. We observe that the channel-aware strategy performs better initially, quickly learning a reasonably good model, but its performance drops sharply after a certain number of iterations and eventually records a large generalization gap compared to the simple random scheduling strategy. This behaviour is due to the biased updates since some devices, those closer to the access point, participate in the federated averaging updates more frequently since they have better channel states on average.\\

\subsubsection{Age-aware Device Scheduling}

As evident from this example, since only a certain portion of the devices are selected for model update in FL, the selection strategy may cause some devices to participate for frequently than others, resulting in biased model updates. Hence, it is important to regulate the frequency of devices' participation in the learning process to prevent any dominance of particular set of devices on the model update. To this end, in \cite{FL.CS-RA4}, an {\em age} based regulation strategy is introduced, where the term `age' is used as a metric to represent the  number of iterations since the last scheduling of a device. Hence, age of the $i$th device, denoted by $a_{i}$, evolves over the iterations in the following way:
\[
    a_{i}(t)=
\begin{cases}
     0, & \text{if } i\in\mathcal{S}_{t}\\
     a_{i}(t)+1, & \text{otherwise}.
\end{cases}
\]
One can observe that the age metric in this context measures the {\em staleness} due to the lack of participation. To this end, the device scheduling strategy can be designed to minimize the impact of the {\em aggregate staleness} at each communication round. For this reason, \cite{FL.CS-RA4} uses the following parameterized function, which guarantees a certain fairness across the devices according to the staleness of their updates:  
\[
    f_{\alpha}(x)=
\begin{cases}
     \frac{x^{1-\alpha}}{1-\alpha}, & \text{if } \alpha\neq1\\
     \log(1+x), & \alpha=1.
\end{cases}
\]
Hence, under certain latency constraint on the communication round, the overall optimization problem can be rewritten as.

\begin{small}
\begin{align}
   \text{\bf P2:}~~~~~~ \;\;\;\min_{\mathbf{P}_{t},\boldsymbol{\Pi}_{t},\boldsymbol{\Phi}_{t}}&
   \begin{aligned}[t]
       \sum_{i\in\mathcal{S}_{t}} f(a_{i,t})  \notag
   \end{aligned} \\
  \text{subject to: }  R_{i,t}(\mathbf{P}_{t},\boldsymbol{\Phi}_{t}) &\geq R_{min}, ~~~ \forall i\in\mathcal{S}_{t},\label{const1}\\ 
 \mathbf{P}_{t} \in \mathcal{P}, & \mathbf{\Phi}_{t} \in \mathcal{F}, \forall t, \label{const2}
\end{align}
\end{small}
where $\mathcal{P}$ and $\mathcal{F}$ denote the feasible power and resource allocation sets, respectively. The first constraint in (\ref{const1}) is introduced to ensure that the duration of each communication round does not exceed a predefined latency constraint $L_{max}=\frac{d}{R_{min}}$. 

The problem {\bf P2} above is fairly general. Consider now a more specific scenario, where we have a set of orthogonal channel resources, i.e., sub-channels, denoted by $\mathcal{W}$ that we allocate to the scheduled devices at each iteration. In this scenario, the feasible sets for power and resource allocation in (\ref{const2}) can be specified to guarantee constraints on the peak and/or average power and the total channel resources, including practical requirements, such as allocating only a single device for each available channel resource. Let $\mathcal{W}_{i,t}$ denote the subset of channel resources allocated to the $i$th device in the $t$th iteration under the resource allocation function $\boldsymbol{\Phi}_{t}$. 

Using Shannon capacity as the transmission rate, which would serve as an upper bound on the rate that can be achieved in practice, and assuming unit-variance additive Gaussian noise at each subchannel, with power allocation $\mathbf{P}_{t}$ and resource allocation $\boldsymbol{\Phi}_{t}$, the transmission rate of the $i$th device is given by
\begin{equation}
R_{i,t}(\mathbf{P}_{t},\boldsymbol{\Phi}_{t})=\sum_{n\in\mathcal{W}_{i,t}} \frac{1}{2}\log(1+G_{i,t,n}P_{i,t,n})
\end{equation}
where $G_{i, t,n}$ denotes the channel gain experienced by the $i$th device in the $t$th iteration over subchannel $n$, and $P_{i,t,n}$ is the power allocated to this subchannel. 

We can impose a power constraint on each device, such that the total power consumption of device $i$ does not exceed $P_{\text{max}}$, i.e.,
\begin{equation}
\sum_{n\in\mathcal{W}_{i,t}} P_{i,n} \leq P_{\text{max}}.
\end{equation}
For the resource allocation, we can require orthogonal transmission to prevent any interference between the devices; and hence, $\mathcal{F}$ is the set of all possible orthogonal resource allocation policies, i.e.,
\begin{equation}
\mathcal{W}_{i,t} \subseteq \mathcal{W} , \forall i,t,~~ \mbox{ and }~~ \mathcal{W}_{i,t} \cap \mathcal{W}_{j,t} = \emptyset, \forall i,j\in\mathcal{S}_{t}.
\end{equation}

Under orthogonal channel allocation, the optimization problem illustrated in \text{\bf P2} is a combinatorial one; hence in \cite{FL.CS-RA4} a greedy heuristic is designed to solve it. The greedy algorithm consists of two phases, executed subsequently.  Given the set of available sub-channels $\widetilde{W}$, the first algorithm solves a resource and power allocation problem for for each device that has not been scheduled yet; that is, for $i\in\mathcal{S}\setminus\mathcal{S}_{t}$, where initially $\mathcal{S}_{t}=\emptyset$, such that the number of channels required for each device is minimized. This is equivalent to solving the following optimization problem for each device $i\in\mathcal{S}\setminus\mathcal{S}_{t}$:

\begin{small}
\begin{align}
   \text{\bf P3:} & \;\;\;\min_{\mathbf{P}_{i,t},\boldsymbol{\Phi}_{i,t}}
   \begin{aligned}[t]
       \vert\mathcal{W}_{i, t} \vert  \notag
   \end{aligned} \\
  \text{subject to: } & R_{i,t}(\mathbf{P}_{i,t},\boldsymbol{\Phi}_{i,t}) \geq R_{min}, \label{const2a}\\ 
 & \sum_{n\in\mathcal{W}_{i,t}} P_{i,t,n} \leq P_{\text{max}}. \label{const2b}
\end{align}
\end{small}

The results of $\text{\bf P3}$ are fed to the second algorithm, which checks the ratio between the impact of the staleness and the number of required sub-channels for the model update, i.e.,
\begin{equation}\label{select_param}
\frac{f_{\alpha}(a_{i,t})}{\vert\mathcal{W}_{i,t} \vert},
\end{equation}
and the device with the highest ratio is added to the set of scheduled devices, and the subchannels identified by $\text{\bf P3}$ for that device are removed from the set of available sub-channels i.e.,
\begin{equation}\label{select_param2}
\mathcal{S}_{t} \leftarrow \mathcal{S}_{t}\cup\left\{i\right\} \text{ and } \widetilde{W} \leftarrow \widetilde{W} \setminus\mathcal{W}_{i,t}.
\end{equation}
Both algorithms are executed subsequently until the remaining sub-channels are not sufficient to add any other device to the set $\mathcal{S}_{t}$, which means there is no feasible solution to the optimization problem \text{\bf P3}.

\subsubsection{Device Scheduling with Fixed Power Allocation}

In general, it is hard to provide a tractable solution to the optimization problem in (\ref{opt1}). Instead, one can assume fixed power transmission, and focus only on the impact of device scheduling, considering many PSs in parallel potentially interfering with each other. We can then replace the communication latency term with the \textit{update success probability} to characterize the convergence performance \cite{FL.CS-RA5}. To be more precise, consider multiple PSs distributed according Poisson point process (PPP) with parameter $\lambda$, which represents the density of the PSs. It is assumed that, in each cluster formed around a PS, there are $N$ devices distributed uniformly, and $K<N$ subchannels are available to be allocated to the devices at each iteration. The channel between the devices and the corresponding PS is assumed to behave according to the {\em block fading propagation} model, in which the channel state remains fixed within a communication block and changes in an independent and identically distributed (i.i.d.) fashion from one transmission block to the next. 

According to the block fading propagation model together with the large scale path-loss model, the received signal-to-interference-plus-noise ratio (SINR) at the PS for device $n$ can be written as
\begin{equation}\label{eqn:snr}
\gamma_{n,t}=\frac{p h_n d_n^{-\alpha}}{\sum_{c\in\mathcal{C}_{outer}}p h_c d_c^{-\alpha} + \sigma^{2}}
\end{equation}
where $p$ is the fixed transmit power for devices, $d_n$ is the distance of device $n$ to the PS, $\sigma^{2}$ is the variance of the additive noise term, $\alpha$ is the path-loss exponent, $h_{n}$ is the small scale fading coefficient, and $\mathcal{C}_{outer}$ is the set of interfering devices belonging to other clusters. Under the given SINR model, a transmission from device $n$ to the PS is considered successful if $\gamma_{n,t}>\gamma^{\star}$ for some predefined threshold $\gamma^{\star}$. Accordingly, the update success probability for device $n$, denoted by $U_{n}$, is defined as the probability of being scheduled for the model update and completing the transmission successfully:
\begin{equation}
U_n=\mathbb{P}(\gamma_{n,t}>\gamma^{\star},n\in\mathcal{S}_{t}).
\end{equation}

Assuming i.i.d. channel statistics across devices within the same clusters and law of large numbers the update success probability can be written as 
\begin{equation}
U_{n} = \lim_{t\rightarrow\infty}\frac{1}{Nt}\sum^{t}_{\tau=0}\sum^{N}_{n=1}\mathbbm{1}_{\left\{n\in\mathcal{S}_{\tau},\gamma_{n,\tau}>\gamma^{\star}\right\}}
\end{equation}
It has been shown in \cite{FL.CS-RA5} that the number of communication rounds to achieve a certain accuracy level can be written as a function of $U_{n}$. More precisely, it is shown to be proportional to $\frac{1}{\log(1-U_{n})}$. The key design question is to find the scheduling policy $\boldsymbol{\Pi}$ that maximizes the update success probability in order to achieve a better convergence result. Three different scheduling policies are investigated in \cite{FL.CS-RA5} for this purpose; namely, random scheduling (RS), round robin scheduling (RR), and proportional fair scheduling (PF).

RS selects $K$ devices randomly for the model update. Under RS, $U_{n}$ can be approximated as
\begin{equation}\label{eqn:usp}
U_n \approx\frac{K/N}{1+V(\gamma^{\star},\alpha)},
\end{equation}
where
\begin{equation}
V(\gamma^{\star},\alpha)=\frac{\sigma^{2}\gamma^{\star}\lambda^{1-\frac{\alpha}{2}}}{P2^{\alpha-2}}(\gamma^{\star})^{\alpha/2}\int^{\infty}_{u=0}\frac{1-\epsilon^{-\frac{12}{5\pi}(\gamma^{\star})^{\alpha/2}u}}{1+u^{\alpha/2}}.
\end{equation}
 Accordingly the minimum number of required iterations under RS, $T_{RS}$, is proportional to 
\begin{equation}
\frac{1}{\log\left(1-\frac{K/N}{1+V(\gamma^{\star},\alpha)} \right)}.
\end{equation}

RR policy, on the other hand, divides the devices into $G=K/N$ groups, and sequentially schedules one group at a time. Compared to RS, the RR policy introduces some level of fairness by giving a chance to each device to contribute to the model update periodically. Under the RR policy the update success probability can be written as 
\begin{equation}
    U_{n}\approx
\begin{cases}
   \frac{1}{1+V(\gamma^{\star},\alpha)} , & \text{if scheduled}\\
   0 & \text{otherwise},
\end{cases}
\end{equation}
and accordingly the minimum number of required iterations under RR policy, $T_{RR}$ is proportional to 
\begin{equation}
\frac{K/N}{\log\left(1-\frac{1}{1+V(\gamma^{\star},\alpha)} \right)}.
\end{equation}

Finally, the PF policy selects the devices opportunistically based on the channel statistics. It sorts all the devices according to the ratio, $\frac{\tilde{\rho}_{i}}{\bar{\rho}_{i}}$,
where $\tilde{\rho}_{i}$ and $\bar{\rho}_{i}$ are the instantaneous signal-to-noise  ratio (SNR) and time-averaged SNR, respectively. The opportunistic behaviour behind the PF policy helps to increase the update success probability, which can be written as 
\begin{equation}\label{eqn:usp}
U_n \approx \sum^{N-K+1}_{i=1} {N-K+1\choose i}\frac{(-1)^{i+1}N/K}{1+V(i\gamma^{\star},\alpha)},
\end{equation}
and the minimum number of required iterations under the RR policy, $T_{PF}$, is proportional to 
\begin{equation}
\frac{1}{\log\left(1-\sum^{K-N+1}_{i=1} {K-N+1\choose i}\frac{(-1)^{i+1}K/N}{1+V(i\gamma^{\star},\alpha)} \right)}.
\end{equation}

To compare the performance of these three scheduling strategies, RR, RS and PF, a classification problem on MNIST data consisting of handwritten numerals, using a convolutional neural network (CNN) architecture has been considered in \cite{FL.CS-RA5}. In the simulations two different scenarios with high and low SINR threshold regimes with $\gamma^{\star}=20dB$ and $\gamma^{\star}=-25dB$, respectively, have been analyzed. The results indicate that, in the high SINR threshold regime, PF significantly outperforms RR policy; while PF achieves $\%94$ average test accuracy, RR gets stuck around $\%50$. This is mainly thanks to the fact that PF achieves further successful global aggregation rounds compared to RR. In the low SINR threshold regime, all three scheduling policies exhibit similar performance since the chance of successfully participating in global aggregation increases with reduced threshold. Further details on the numerical experiments can be found in \cite{FL.CS-RA5}.

In (\ref{opt1}), we assumed a fixed number of scheduled devices at each iteration, and tried to minimize the latency of getting updates from so many devices. Alternatively, we can impose a deadline on each iteration, and try to receive updates from as many devices as possible at each iteration \cite{FL.CS-RA1}. Assuming that the devices transmit one-by-one using all the available channel resources, the device selection problem can be formulated as follows:

\begin{small}\label{opt2}
\begin{align}
   \text{\bf P4:} \;\;\;&\max 
   \begin{aligned}[t]
       \vert\mathcal{S}_{t}\vert  \notag
   \end{aligned} \\
  &\text{subject to: }   \mathcal{S}_{t} \subseteq \bar{\mathcal{S}_{t}}, ~~~ \forall t, \\
  &\sum^{\vert\mathcal{S}_{t}\vert}_{i=1} \left( L^{\text{comm}}_{s_i,t}  + \max\left(\sum^{\vert\mathcal{S}_{t}\vert}_{i=1} L^{\text{comm}}_{s_i,t}, L^{\text{\text{comp}}}_{s_i,t} \right) \right)  \leq T_{max}, \label{latency2}
\end{align}
\end{small}
where the objective is to schedule the maximum number devices from among the randomly sampled set $\bar{\mathcal{S}_{t}}$, while the overall latency is bounded by some predefined  $T_{max}$. 
We want to highlight that, in this formulation, the device selection strategy also incorporates the scheduling decision for the model updates, such that the devices in set $\mathcal{S}_{t}=\left\{s_{1},\ldots,s_{\vert\mathcal{S}_{t}\vert}\right\}$ are ordered. This ordering also affects the latency since, as illustrated in (\ref{latency2}), the computation time of the $s_{i}$th device can be overlapped with the cumulative latency of the previously scheduled devices. In general, $\mathbf{P4}$ is a combinatorial problem, thus a greedy strategy is proposed in \cite{FL.CS-RA1}, similar to those used for the knapsack problem. This strategy iteratively adds to set $\mathcal{S}_{t}$ the device that introduces the minimum additional delay, until the cumulative latency reaches $T_{max}$.

Latency-aware device selection problem with a fairness constraint is also studied in \cite{FL.CS-RA3}, where the optimal device selection problem is analyzed as a multi-armed bandit (MAB) problem.\\

\subsubsection{Update-aware Device Scheduling}

In all the above policies, the devices are scheduled according to the channel conditions, with the additional consideration of the age or the fairness in order to make sure that each device is scheduled at some minimal frequency. An alternative approach to these policies is considered in \cite{amiri:arxiv:20}, where the updates of the devices are also taken into account when selecting the devices to be scheduled at each iteration. 

Assume that we schedule a fixed number of $K$ devices at each iteration. Differently from the previous formulation, here we fix both the latency and the number of devices at each iteration, but instead change the update resolution sent from each device, which was set to $d$ bits in (\ref{opt1}). Once the scheduled devices are decided, the channel resources are allocated between these devices such that each device can approximately send the same number of bits to the PS. 

If we schedule the devices with the best channel states, the scheduled devices will be able to transmit their model updates with the highest fidelity; however, not all model updates may have the same utility for the SGD process. For example, if the data at one of the devices is already in accordance with the current model, this device will not want to change the model, and hence, will send a very small or zero model update to the PS, which can be ignored without much loss in the optimization. To enable such an ``update-aware'' scheduling policy, \cite{amiri:arxiv:20} proposes to schedule the devices taking into account the $l_2$ norm of their model updates. 

Four different scheduling policies are proposed and compared in \cite{amiri:arxiv:20}: the \textit{best channel} (BC) policy schedules devices soleley based on their channel conditions, i.e., selects the $K$ devices with the best channel conditions, \textit{best $l_2$-norm} (BN2) strategy schedules the $K$ devices with the best channel conditions. On the other hand, \textit{best channel-best $l_2$-norm} (BC-BN2) strategy first chooses $K_c$ devices according to their channel conditions, and then chooses $K$ out of these $K_c$ devices, depending on the $l_2$ norm of their updates. This guarantees scheduling users with both good channel states and significant model updates. Note however that, the real quality of the update transmitted to the PS depends also on the available channel resources, which determines how much the device needs to quantize its model update. For example, a device might have an update with a relatively large $l_2$ norm; however, this might reduce drastically after quantization if the channel of this device is relatively weak. To take into account this reduction in the update quality due to quantization, the \textit{best $l_2$-norm-channel} (BN2-C) strategy chooses the final devices based on the $l_2$ norm of the updates they would send if they were the sole transmitter.

\begin{figure}[t!]
\centering
\centering
\includegraphics[scale=0.8,trim={20pt 7pt 45pt 40pt},clip]{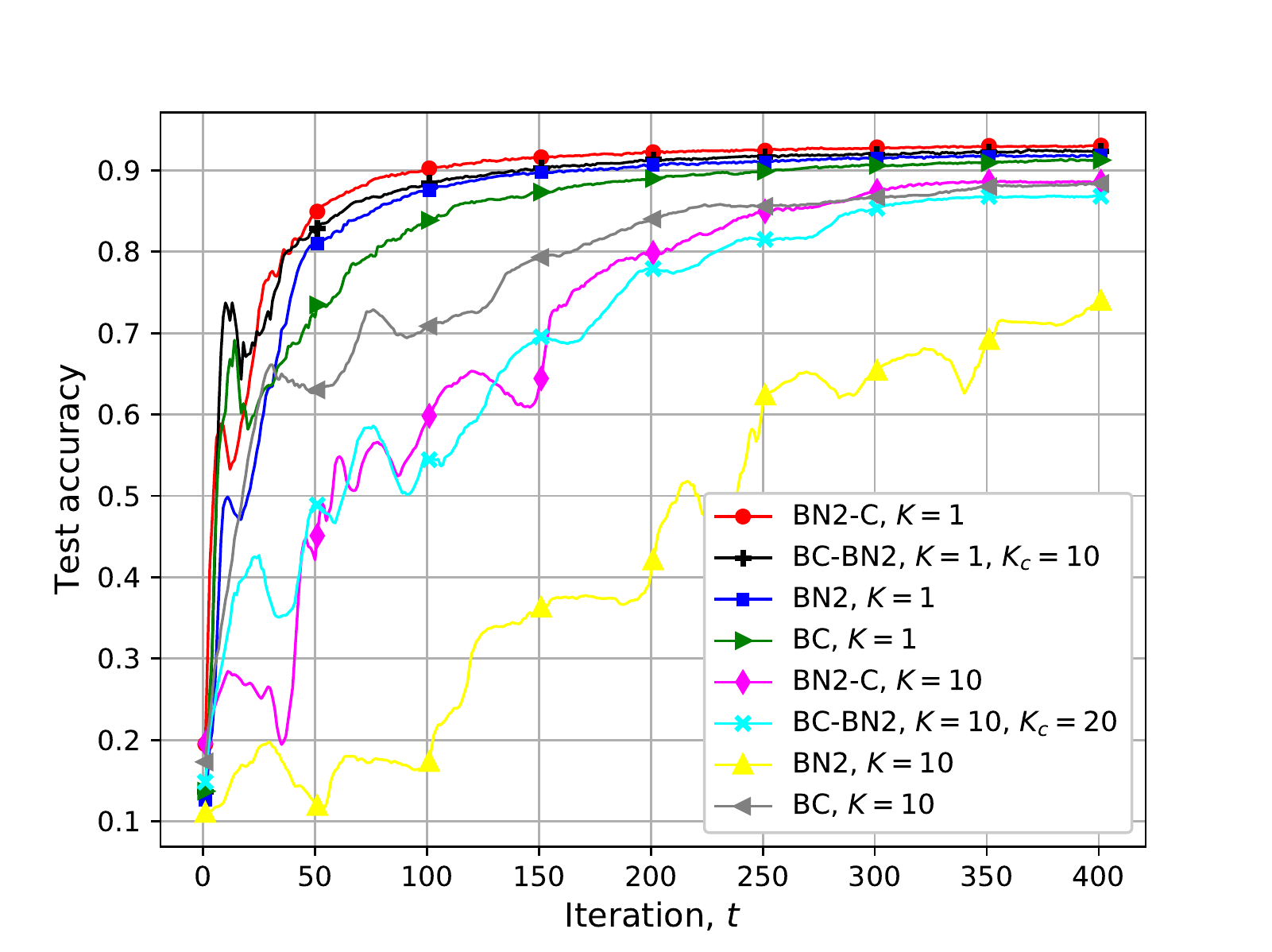}
\caption{Performance of different scheduling policies with $N=40$ devices.}
\label{Fig_IID_K_1_5}
\end{figure}

We compare the test accuracy of these four schemes in Fig. \ref{Fig_IID_K_1_5} considering the training of a neural network model for MNIST image classification. We considering 40 devices, each with 1000 randomly chosen MNIST samples. We can see from this plot that scheduling only a single device at each iteration achieves the best performance for all four schemes. We can also see that scheduling solely based on the channel gains or on the update significance has a slower convergence speed, and converges to a lower test accuracy. It is clear that the best performance is achieved when both the channel gain and the update significance is taken into account when scheduling the devices.

\subsection{Hierarchical Edge Learning}

In general, to prevent excessive communication load and congestion at the PS only a small portion of devices are scheduled at each round to upload their locally trained models. On the other hand, scheduling only a small number of $K$ devices at each round increases the required number of communication rounds to achieve a target accuracy level. In order to reduce the communication load while keeping the number of communication rounds fixed, orchestration of the edge learning framework can be modified such that multiple PSs are employed to utilize parallel aggregation in the network under the orchestration of a main coordinator PS \cite{HFL1, HFL2}.

\begin{algorithm}[ht!]
\caption{Hierarchical Federated Learning (HFL)}\label{alg:hfl}
\begin{algorithmic}[1]
\For{$t=1,2,\ldots$}
    \For{l=1,\ldots,L} in parallel
        \For{$i\in \mathcal{C}_{l}$} in parallel
        \State Pull $\boldsymbol{\theta}_{l,t-1}$ from $PS_{l}$: $\boldsymbol{\theta}^{0}_{i,t-1}=\boldsymbol{\theta}_{l,t-1}$
            \For{$h=1,\ldots,H_{local}$} 
            \State Compute SGD: $\mathbf{g}_{i,h}=\nabla_{\boldsymbol{\theta}}\mathcal{L}_{n}(\boldsymbol{\theta}^{h}_{i,t-1},\zeta_{i,h})$
            \State Update model: $\boldsymbol{\theta}^{h}_{i,t-1}=\boldsymbol{\theta}^{h-1}_{i,t-1}-\eta_{t}\mathbf{g}_{i,h}$
            \EndFor
            \State Push $\boldsymbol{\theta}^{H_{local}}_{i,t-1}$ to $PS_{l}$
        \EndFor
    \State{\textbf{ Intra-cluster averaging}:} \State $\boldsymbol{\theta}_{l,t}=\frac{1}{ \vert \mathcal{C}_{l} \vert}\sum_{i\in\mathcal{C}_{l}} \boldsymbol{\theta}^{H_{local}}_{i,t-1}$
    \EndFor
    \If{$t \vert H$}:
    \State \textbf{Inter-cluster Averaging}
    \State $\boldsymbol{\theta}_{t}=\frac{1}{L}\sum^{L}_{l=1} \boldsymbol{\theta}_{l,t}$
    \EndIf
\EndFor
\end{algorithmic}
\end{algorithm}

The generic structure of the {\em hierarchical FL (HFL)} framework is provided in Algorithm \ref{alg:hfl}. In HFL, devices in the network are grouped into $L$ clusters and the devices within the same cluster are assigned to the same PS. Therefore, intra-cluster federated averaging is executed in parallel (line 2-10 in Algorithm \ref{alg:hfl}). Following $H$ rounds of intra-cluster averaging, PSs communicate with each other through the main PS to seek a consensus on the global model (line 13 in Algorithm \ref{alg:hfl}).

\begin{figure*}[t]
  \centering
  \includegraphics[width=1\linewidth]{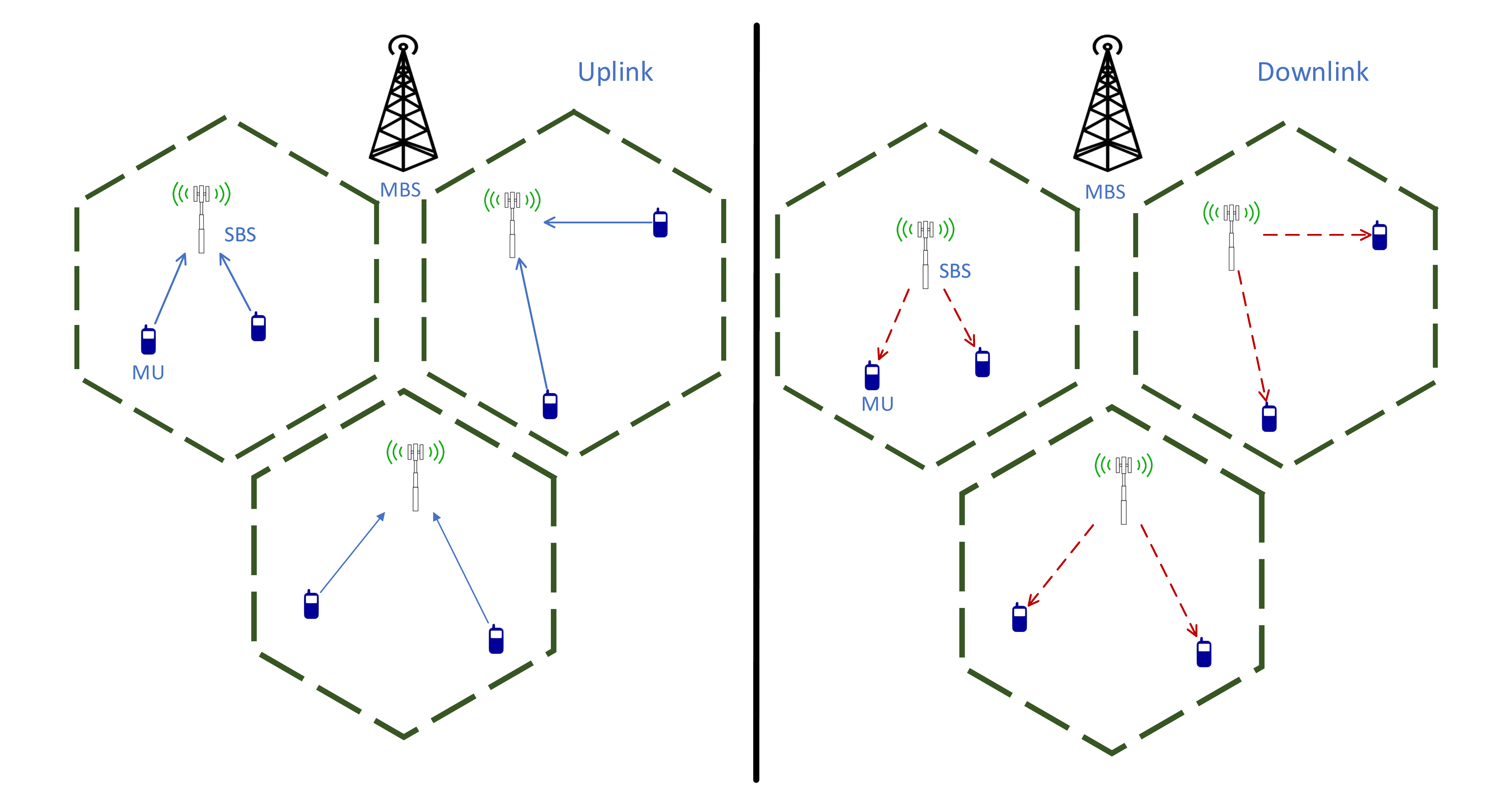}  
  \caption{Intra cluster consensus.}
  \label{fig:HFLl}
\end{figure*}

\begin{figure*}[ht]
  \centering
  \includegraphics[width=1\linewidth]{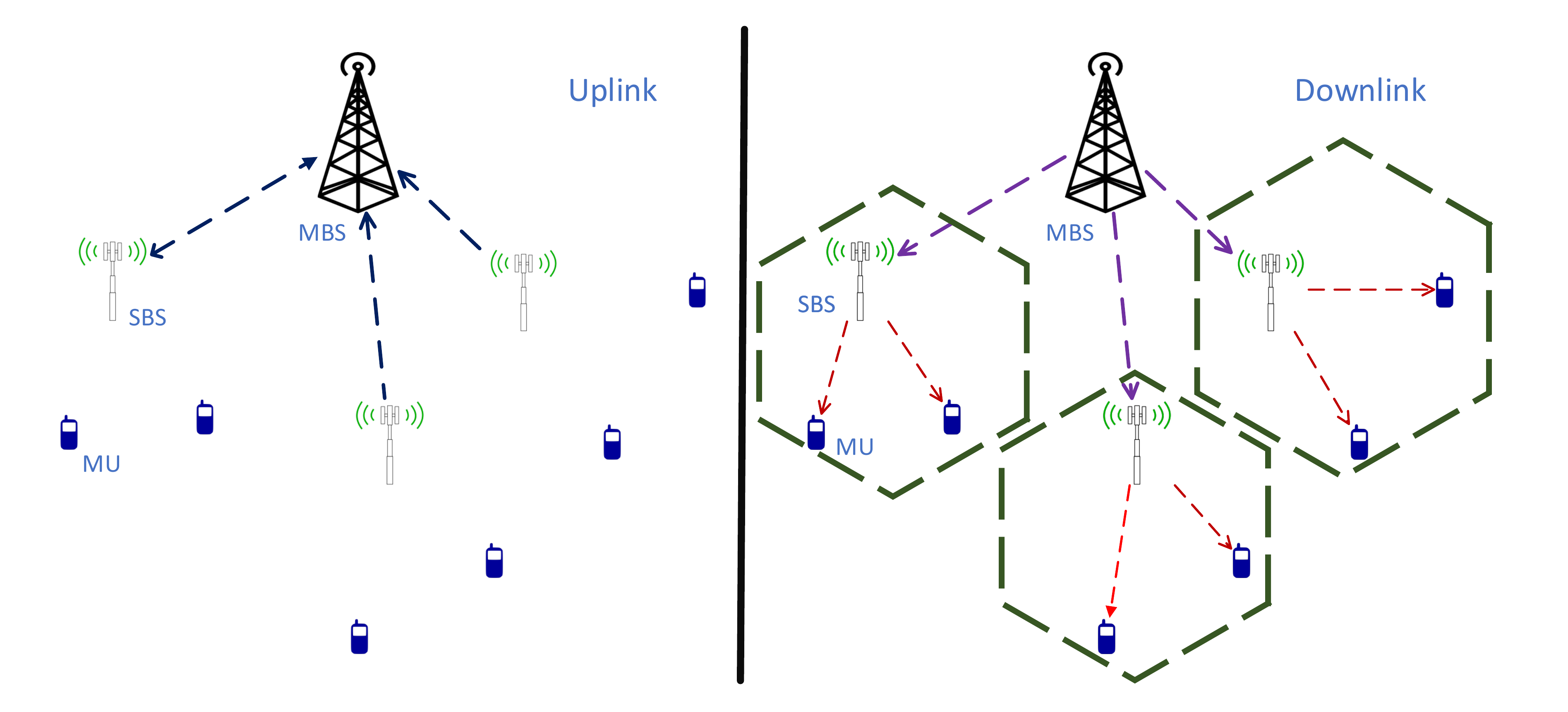} 
  \caption{Inter cluster consensus.}
  \label{fig:HFLg}
\label{fig:HFL}
\end{figure*}

In the wireless setting, this hierarchical edge learning strategy can be implemented through small-cell base stations (SBS) such that the clusters are formed by assigning devices to SBSs and the SBSs are orchestrated by a macro-cell base station (MBS), as illustrated in Fig. \ref{fig:HFLl} and  Fig. \ref{fig:HFLg}. In this scenario, we refer to the devices as mobile users (MUs).


To illustrate the benefits of the hierarchical FL strategy, we consider $28$ MUs uniformly distributed in a circular area with radius $750$ meters. We consider hexagonal clusters, where the diameter of a circle inscribed in each of them is $500$ meters. We consider in total $7$ clusters with hexagonal shape and the SBSs are located at the center of the hexagons. 
We further assume that the fronthaul links between the MBS and the SBSs are $100$ times faster than the uplink and downlink connections between the MUs and the SBSs. We assume $600$ subcarriers with subcarrier spacing of $30$ KHz. The maximum transmit powers of the MBS, SBSs, and the MUs are set to $20$W, $6.3$W, and $0.2$W, respectively.


For the simulations, we consider the image classification problem over the CIFAR-10 dataset with 10 different image classes and collaboratively train the  ResNet18  architecture. The dataset is divided disjointly among the devices in an i.i.d. manner. We set the batch size for training to $64$ and the initial learning rate to $0.25$, which is reduced by a factor of $10$ at the end of the 150th epoch and 225th epoch. Finally, we apply 99\%, 90\%, 90\% and 90\% sparsification for communication from the MUs to the SBSs, SBSs to MUs, SBSs to the MBS, and the MBS to SBSs, respectively. We set $H_{local}=1$ in Algorithm \ref{alg:hfl}, and use the \emph{momentum} optimizer for the local updates.

\begin{figure}
\centering
     \includegraphics[scale=0.6]{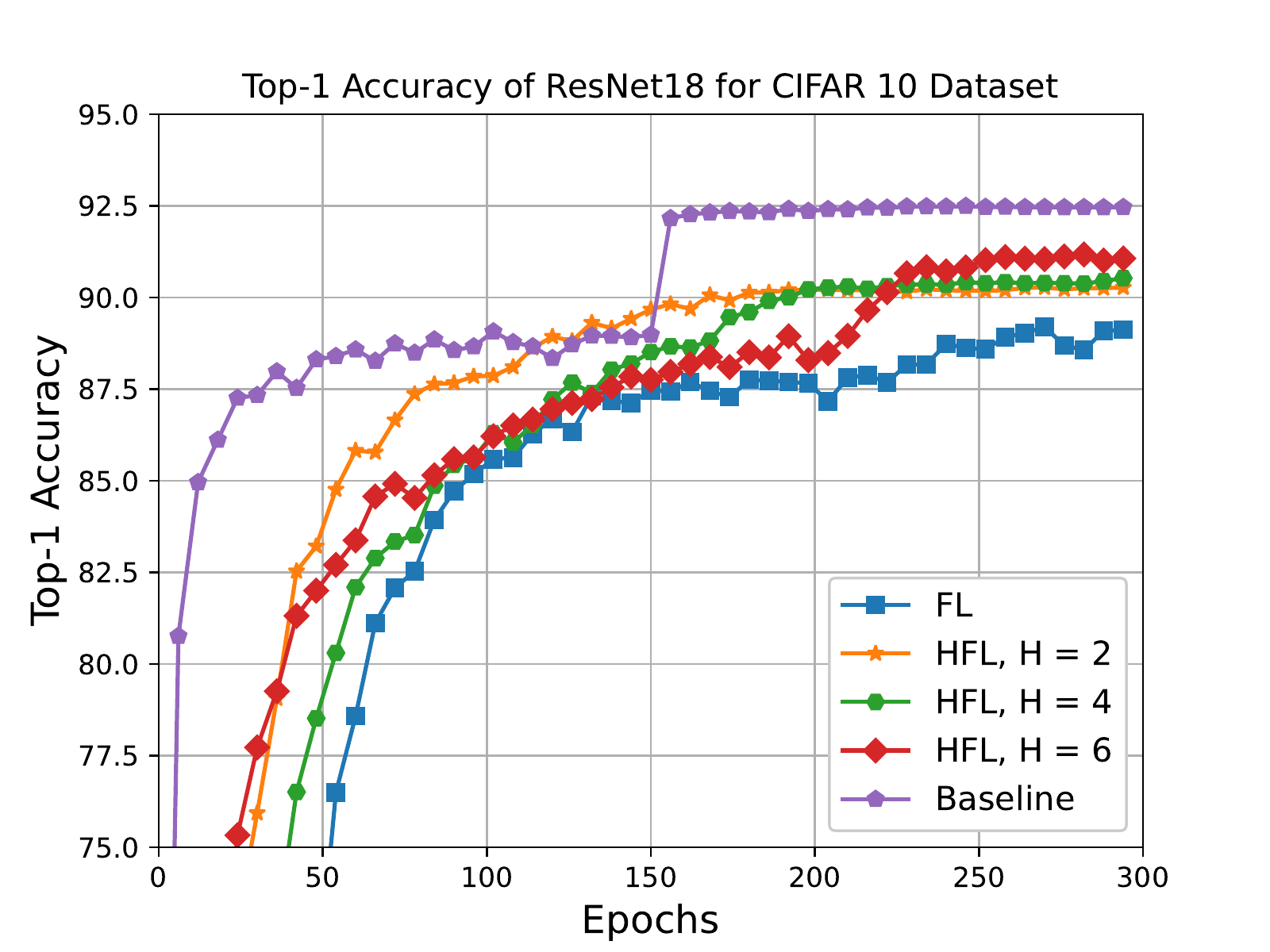}
				\caption{Test accuracy through iterations.}
		\label{fig:resnetacc}
\end{figure}

The convergence results for the HFL framework for parameters $H=2,4,6$, as well as the centralized FL and the baseline method corresponding to centralized single machine training are illustrated in Fig. \ref{fig:resnetacc}, while the final test accuracy results are presented in Table \ref{tab:acc}. We observe that HFL achieves a higher final accuracy as well as faster convergence compared to conventional FL. Additionally, the hierarchical FL strategy provides 5 to 7 times speed up in the latency required to reach these accuracy levels compared to FL orchestrated directly by the MBS.

\begin{table}
 \caption{Test accuracy results for different strategies.}
 \label{tab:acc}
\begin{center}
    \begin{tabular}{ |c |c|}
    \hline
 		   Baseline  & $92.48\pm 0.13$\\ \hline 
          FL & $89.23\pm 0.42$\\ \hline 
          HFL, $H=2$ & $90.27\pm 0.11$\\ \hline 
          HFL, $H=4$ & $90.474\pm 0.20$\\ \hline 
          HFL, $H=6$ & $91.03\pm 0.19$\\ \hline 
     \end{tabular}		
 \end{center}
\end{table}

\section{Summary}

In this chapter, we presented a brief introduction to distributed/ collaborative learning, particularly targeting  wireless edge applications. We first introduced general communication reduction methods such as sparsification, quantization, and local iterations. We then provided an overview of device scheduling and resource allocation strategies for distributed learning over wireless networks. We particularly highlighted the fact that both the scheduling and the resource allocation strategy can be substantially different from conventional solutions for wireless network optimization, which typically aims at maximizing the throughput, or minimizing the delay. In distributed learning, the goal is to solve the underlying optimization problem as fast and as accurately as possible, and we often employ iterative algorithms, such as SGD. This means that, in addition to receiving accurate solutions from the participating devices, we also need to guarantee that each device is scheduled with some minimal regularity. Additionally, selecting the devices depending on the significance of their updates on the solution can increase the convergence to the optimal solution. 

There are many other factors that can be considered in formulating the optimal scheduling scheme, such as the energy consumption at the devices \cite{zeng2020energyefficient}, or impact of the compression in the PS-to-device links \cite{amiri2020convergence}. We also remark that we focused exclusively on digital schemes in this chapter, where the devices first compress their model updates into a finite number of bits, and transmit these bits to the PS over orthogonal channels. An alternative approach is to transmit all the model updates from the devices simultaneously in an uncoded/ analog manner, which allows exploiting the signal superposition property of the wireless channel for over-the-air computation \cite{Amiri:TSP:20, Amiri:TWC:20}.
\bibliographystyle{IEEEtran}
\bibliography{IEEEabrv,ref.bib}

\ifCLASSOPTIONcaptionsoff
  \newpage
\fi




\end{document}